\documentclass[conference]{IEEEtran}
\IEEEoverridecommandlockouts
\usepackage{cite}
\usepackage{amsmath,amssymb,amsfonts}
\usepackage{algorithmic}
\usepackage{algorithm}
\usepackage{graphicx}
\usepackage{textcomp}
\usepackage{xcolor}
\usepackage{url}
\def\BibTeX{{\rm B\kern-.05em{\sc i\kern-.025em b}\kern-.08em
    T\kern-.1667em\lower.7ex\hbox{E}\kern-.125emX}}
\begin{document}

\title{Histogram-Based Federated XGBoost using Minimal Variance Sampling for Federated Tabular Data}

\author{\IEEEauthorblockN{1\textsuperscript{st} William Lindskog}
\IEEEauthorblockA{\textit{Corporate Research and Development} \\
\textit{DENSO Automotive Deutschland GmbH}\\
Munich, Germany \\
w.lindskog@eu.denso.com}
\and
\IEEEauthorblockN{2\textsuperscript{nd} Christian Prehofer}
\IEEEauthorblockA{\textit{Corporate Research and Development} \\
\textit{DENSO Automotive Deutschland GmbH}\\
Munich, Germany \\
c.prehofer@eu.denso.com}
\and
\IEEEauthorblockN{3\textsuperscript{rd} Sarandeep Singh}
\IEEEauthorblockA{\textit{Corporate Research and Development} \\
\textit{DENSO Automotive Deutschland GmbH}\\
Munich, Germany \\
s.singh.1@eu.denso.com}
}

\maketitle

\begin{abstract}
    Federated Learning (FL) has gained considerable traction, yet, for tabular data, FL has received less attention. Most FL research has focused on Neural Networks while Tree-Based Models (TBMs) such as XGBoost have historically performed better on tabular data. It has been shown that subsampling of training data when building trees can improve performance but it is an open problem whether such subsampling can improve performance in FL. In this paper, we evaluate a histogram-based federated XGBoost that uses Minimal Variance Sampling (MVS). We demonstrate the underlying algorithm and show that our model using MVS can improve performance in terms of accuracy and regression error in a federated setting. In our evaluation, our model using MVS performs better than uniform (random) sampling and no sampling at all. It achieves both outstanding local and global performance on a new set of federated tabular datasets. Federated XGBoost using MVS also outperforms centralized XGBoost in half of the studied cases. 
\end{abstract}


\section{Introduction}\label{sec:introduction}
Federated Learning (FL) \cite{konevcny2016federated} has gained traction, mainly due to enhanced privacy and ability to exploit distributed datasets. Generally, FL offers low communication cost , privacy aware, Machine Learning (ML) functions and research suggest that its performance can be similar to that of centralized ML \cite{zhang2021survey}.  

FL was initially designed to fit \textit{parametric} models, models characterized by a simplified input-output function of a know form e.g. (Deep) Neural Networks (DNNs). DNNs show great promise but struggle with tabular data, a common data type \cite{borisov2022deep}. Researchers have designed parametric models for tabular data \cite{arik2021tabnet}, suitable for FL \cite{lindskog2022federated}, yet they do not provide state-of-the-art performance \cite{shwartz2022tabular}. \textit{Non-parametric} models have performed well on tabular data. They do not make strong assumptions about the form of the input-output function and can learn the form in training. Tree-Based Models (TBMs) e.g. Random Forests and boosted trees \cite{kern2019tree} have shown great performance on tabular data. However, federated implementation is non-trivial and researchers have only recently designed federated TBMs.
\begin{figure}[t]
    \centering
    \includegraphics[width=\linewidth]{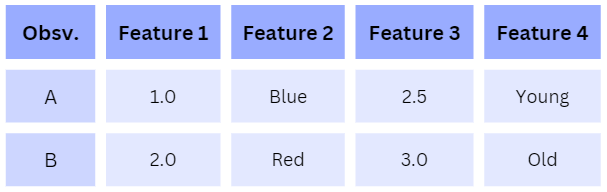}
    \caption{Tabular data, each row is a unique observation and the columns indicate features. Values can be numerical and categorical.}
    \label{fig:tabular_data}
\end{figure}
Extreme Gradient Boosting (XGBoost) \cite{chen2015xgboost} can be considered a suitable model for such data, often outperforming related models \cite{borisov2022deep}, yet few XGBoosts for FL are available. 

Recent work has shown that subsampling for TBMs can improve performance in a centralized ML \cite{ibragimov2019minimal}. Subsampling is the process of taking a fraction of data to train on. It can help select useful data. We refer to subsampling as \textit{sampling}. Sampling has received little attention in FL and initial works do not show increase in performance, compared to centralized learning \cite{chen2021secureboost+}. In FL, each client gets a subset of the whole dataset for training, and with additional sampling we further reduce the local datasets with different techniques. 

To such end, we propose histogram-based Federated Learning with XGBoost using sampling of training data when building the trees. We evaluate our model using uniform sampling and Minimal Variance Sampling (MVS) \cite{ibragimov2019minimal}, as MVS shows performance improvements in centralized settings. We do not only study the global performance but also local performance on given client test sets. We also present a set of federated tabular datasets that reflect the natural properties of data in FL. We focus on developing models for horizontal FL and from our extensive evaluation, our main contributions are:
\begin{itemize}
    \item Federated XGBoost using MVS improves performance in terms of accuracy and regression error when compared with federated XGBoost using no- or uniform sampling. 
    \item Federated XGBoost using MVS performs similarly as centralized, and even outperforms it in half of the cases.
    \item FedTab - A selected set of federated tabular datasets that can serve future benchmark studies.
\end{itemize}

The paper is structured as related work, method and results before a discussion and conclusion. 
\section{Related Work}\label{sec:related_work}
Tabular data are observations that are represented by rows and columns, see Figure \ref{fig:tabular_data}. FL was initially designed for parametric models yet research suggests that TBMs perform better on tabular data \cite{shwartz2022tabular}. Researchers have started to develop federated TBMs and few provide sophisticated open-source packages: FATE \cite{fate_2022}, Flower \cite{beutel2020flower} and NVFlare \cite{roth2022nvidia}. 

In their paper, \cite{liu2020federated} present a non-parametric federated representation of a random forest. Building their federated forest requires clients to train partial tree models based on a subsets of features and data, and aggregation of these on server side. Other researchers have thereafter implemented their versions of federated random forests, focusing on various aspects such as decentralization \cite{de2020dfedforest} and blockchain-based \cite{aliyu2021blockchain} forests. Yet, the performance of a federated random forest is similar or slightly lower than for centralized learning. In centralized learning, there is reason to believe that the strongest model is still XGBoost \cite{shwartz2022tabular}. In a federated setting, \cite{wu2020privacy} found that one of the stronger performing models are federated Gradient Boosted Decision Trees (GBDTs). Much research has thereafter focused on developing federated GBDTs. 

\subsection{Federated Gradient Boosted Decision Trees}\label{subsec:(RELATED_WORK)_federated_gradient_boosted_decision_trees}
Researchers started to investigate federated GBDTs in 2019. \cite{yang2019tradeoff} and \cite{liu2019boosting} proposed federated GBDTs, mainly for classification tasks. They explore features such as secure aggregation and discuss model efficiency. A similar study was made by \cite{feng2019securegbm} in which they argue that a secure federated LightGBM can be used for tabular data. FederBoost \cite{tian2020federboost} and SecureBoost \cite{cheng2021secureboost} extended their work by introducing more complete federated XGBoost frameworks. The authors of SecureBoost also demonstrate that their model can be as accurate as non-federated TBMs. \cite{fedtree} created FedTree, an open-source framework for federated forests and GBDTs that can be used for horizontal- and vertical FL. \cite{ong2020adaptive} created a federated XGBoost that could adapt to local data. Party Adaptive XGBoost (PAX) includes a quantile sketch approximation hyperparameter $\epsilon$ which is roughly the inverse of \textit{bin size of histogram}. Given a global $\epsilon^{(A)}$,  they define clients' approximation hyperparameter as 

\begin{equation}
        \mathbf{\epsilon_i} = \epsilon^{(A)} \bigg(\frac{|d_i|}{\sum_{d \in D}|d|}\bigg)
        \label{eq:local_eps}
\end{equation}


for which $|d_i|$ is the size of the client dataset $i$, a subset of total dataset $D$. The inverse $1 / \epsilon_{i}$ gives the bin size for each client. Their PAX model has thereafter been used in other studies \cite{jones2022federated} in which it has been applied to non-independent and identically (non-IID) distributions. \cite{chen2021secureboost+} extended SecureBoost to include Gradient One-Side Sampling (GOSS), a sampling technique for selecting training data \cite{ke2017lightgbm}. Sampling training data can improve performance in centralized learning \cite{bentejac2021comparative} yet it is an open problem whether the effect is similar for FL. Their model SecureBoost+ reduced tree building time and provided similar performance to that of XGBoost. \cite{rizk2021optimal} presented \textit{importance sampling} for FL. Their algorithm ISFedAvg not only samples specific data instances from mini-batches but also incorporates a dynamic client selection. \cite{zhu2022isfl} extended their work to address non-IID data. Their ISFL framework improves the original FedAvg algorithm but still does not outperform centralized learning.

In a centralized setting, \cite{ibragimov2019minimal} proposed Minimal Variance Sampling (MVS), a sampling technique that they argue outperforms GOSS, similarly concluded by \cite{ou2020out}. MVS uses the regularized gradients $\hat{g}_i = \sqrt{g^2_i + \lambda h^2_i}$ for which $g_i$ and $h_i$ are the gradients and hessian, to calculate the probability of selecting a specific data point. It outperforms other methods e.g. GOSS on small and large datasets, can improve computation time, and generalize better than other methods. Summarized, MVS bases its sample selection criteria on low variance output from previous predictions and selects a predefined proportion of samples to help grow the tree. This approach is similar to \cite{sokolovska2021vanishing}, with less computation as we do not include any corrective fine-tuning. To such end, we evaluate whether federated XGBoost using MVS can improve performance in terms of accuracy (and other related scores) in a federated setting. 

\section{Method}\label{sec:method}
This section includes basics of XGBoost, our federated implementation, and a set of compiled datasets. The datasets are carefully selected to fit our comparisons and a federated setting. 

\begin{table*}[tb]
    \centering
    \caption{Overview of included datasets. ID column includes number unique IDs that can be used in FL. Insurance dataset has no class since it is a regression dataset. The lower 3 datasets are used in other related work, thus we include and evaluate our model on them. }
    \begin{tabular}{l|r|r|r|r|c|c|c}
    \hline
    \textbf{Dataset} & \textbf{Size} & \textbf{\# Features} & \textbf{\# Class} & \textbf{\# IDs} & \textbf{Split Feature} & \textbf{License} & \textbf{Reference}\\
    \hline
    FEMNIST & 382.705& 784 & 10 & 3.382 & user\_id & BSD-2-Clause & \cite{caldas2018leaf}\\
    Synthetic & 151.152 & 30& 30& 1.000 & User\_ID & BSD-2-Clause & \cite{caldas2018leaf}\\
    Machine Failure & 10.000&6 & 2& 3&Type & CC0 & \cite{matzka2020explainable}\\
    Lumpy Skin & 5.039&18 &2 & 22&country & CC BY 4.0 & \cite{Afshari_Safavi2021-ee}\\
    Heart Disease & 740 & 11& 2& 4 & source & CC BY 4.0 & \cite{Dua:2019}\\
    Insurance &1.338 & 6 & - & 4 & region & CC0 & \cite{Kaggle_Insurance_dataset}\\
    \hline
    Airline Delay & 4.000& 25&2& - & - & CC0 & \cite{Kaggle_Airline_dataset}\\
    Credit Card & 284.807& 29&2 & - & - & DbCL v1.0 & \cite{Kaggle_Credit_dataset}\\
    Firewall & 65.532& 10& 4 & - & - & CC BY 4.0 & \cite{Dua:2019}\\
    \hline
    \end{tabular}
    \label{tab:my_label}
\end{table*}

\subsection{XGBoost Model}\label{subsec:(METHOD)_federated_XGBoost}
The model architecture we opt to use in this study is XGBoost and we hereafter present some preliminaries to the model. Given a dataset $D\in\{x_i, y_i \}^N_{i=1}$ where $x\in\mathbb{R}^d$ with $d$ features, $y\in\mathbb{R}$, and an arbitrary task, predictions are calculated for $K$ trees and $\eta$ learning rate as:

\begin{equation}
    \hat{y}_i = \sum^{K}_{k=1} \eta f_k(x_i) 
    \label{eq:xgboost_prediction}
\end{equation}

where $f_k(x_i)$ is the prediction made by the $k-$th tree. XGBoost's objective is minimizing sum the total loss for all samples and regularization parameter. $L(\phi)$ is training loss and $\Omega$ is regularization term. 

\begin{equation}
    L(\phi) = \sum_il(y_i, \hat{y}_i) + \sum_k\Omega(f_k)
    \label{eq:loss_xgboost}
\end{equation}

The Taylor expansion of the objective function is:

\begin{equation}
    L^{(t)} \simeq \sum^n_{i=1}\Big[l\big(y_i, \hat{y}^{(t-1)}_i\big) + g_if_t + \frac{1}{2}h_if^2_t\Big] + \Omega(f_t)
    \label{eq:loss_xgboost_Taylor}
\end{equation}



for which gradient $g_i$ and hessian $h_i$ are calculated as: 


\begin{equation}
    g_i = \partial_{\hat{y}^{(t-1)}_i} l(y_i, \hat{y}^{(t-1)}_i), h_i = \partial^2_{\hat{y}^{(t-1)}_i)} l(y_i, \hat{y}^{(t-1)}_i)
    \label{eq:gradient_and_hessian}
\end{equation}

for which $\hat{y}^{(t-1)}_i$ is previous prediction made by tree and $l(y_i, \hat{y}^{(t-1)}_i)$ is loss function. 







\begin{figure}[tb]
    \centering
    \includegraphics[width=\columnwidth]{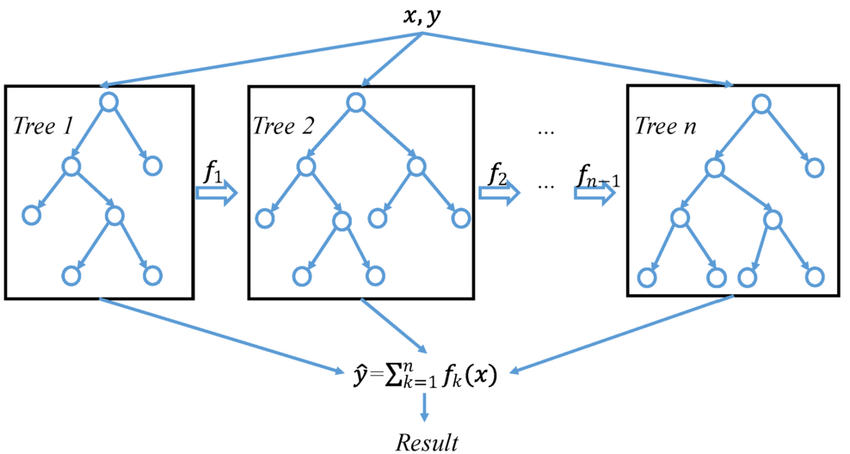}
    \caption{General architecture of XGBoost. XGBoost combines predictions of several individual models, called "weak learners", into final prediction. }
    \label{fig:xgboost}
\end{figure}

In summary, XGBoost works by combining the predictions of several individual models, called "weak learners", into a final prediction. Each weak learner is typically a simple decision tree. XGBoost trains weak learners iteratively and adjusts their predictions based on how well they perform on the data. In each iteration, the algorithm adds a new weak learner to the ensemble and adjusts the weights of the previous learners to account for any mistakes they made, see Figure \ref{fig:xgboost} for reference. Using MVS, the training samples with a higher gradient and hessian are more likely to be selected for training.  

\begin{figure*}[tb]
    \centering
    \includegraphics[width = \linewidth]{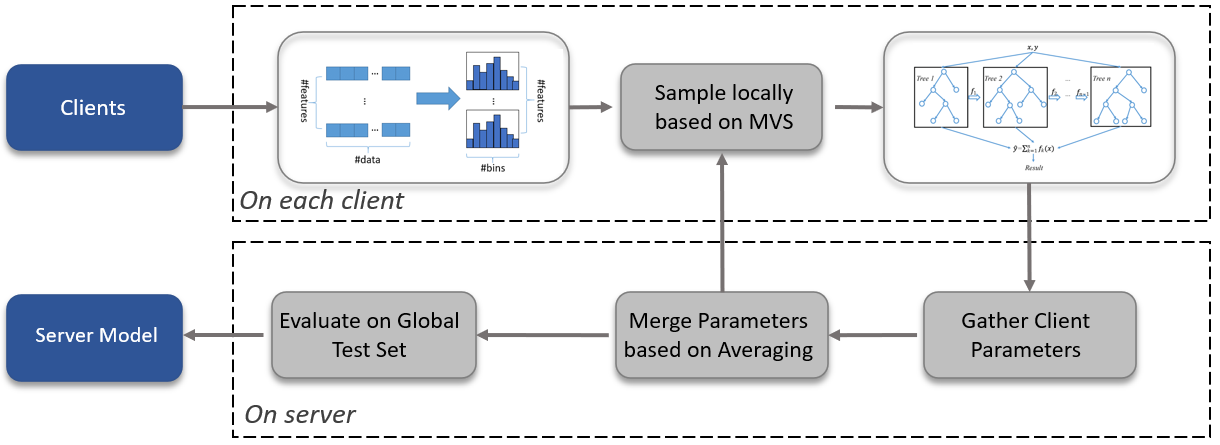}
    \caption{System overview of federated XGBoost using sampling.}
    \label{fig:(METHOD)_xgboost_flow}
\end{figure*}


\subsection{Federated XGBoost using Sampling}\label{subsec:(METHOD)_federated_XGBoost_sampling}
Instead of looking at framework e.g. PAX \cite{ong2020adaptive}, we propose federated XGBoost using MVS. The core part of MVS is calculating the regularized gradients $\hat{g}_i = \sqrt{g^2_i + \lambda h^2_i}$ for which $g_i$ and $h_i$ are the gradient and hessian. MVS bases its selection criteria on low variance output from previous predictions and selects a proportion of samples to grow the tree. Selecting instances with low variance can provide more stable and informative training examples. Moreover, we opt to use histogram-based XGBoost framework as it can improve efficiency and robustness to outliers \cite{chen2015xgboost} even though there might be a loss of information due to approximation. Histograms are constructed based on distribution of feature values. 


The process of building a federated XGBoost for horizontal is explained in Algorithm \ref{alg:flexbo} and we show our contribution to the algorithm. The data are split and distributed to clients (Line 1). Gradient boosted decision trees then use \textit{quantile based approximation} to reduce the search space of finding optimal splits, we use the work proposed in \cite{ou2020out}.
\begin{algorithm}[tb]
    \caption{Federated XGBoost with Sampling}
    \label{alg:flexbo}
    \textbf{Input}: $D$, Data; $A$, Aggregator; $C$, Number of Clients; \\
    \textbf{Parameter}:  $S$, Sampling Fraction; $N$, Number of Training Rounds; $K$, Early Stopping Rounds, $\eta$, Learning Rate; $\lambda$, L2 Regularization; $B$, Max. number of Bins; $T$, Tree Depth\\
    \textbf{Output}: Federated XGBoost Model 
    \hskip-1em\vtop{\vskip.0cm\hsize=3.5in \hrulefill}
    \begin{algorithmic}[1] 
    \STATE $f^{(A)}_{\emptyset} = $ \textit{initialize\_model}($\eta, \lambda, B, T$)
    \STATE
    \FOR{$c \gets 1 \text{ to } C$}
    \STATE $\Tilde{D}^{c_i}_X$ $\gets$ \textit{construct\_histogram}($D^{c_i}_X, B$) 
    \STATE $c_i$ transmits $\Tilde{D}^{c_i}_X$ to $A$
    \ENDFOR
    \STATE $\Tilde{D}^{(A)}_X$ $\gets$ \textit{merge\_histogram}($\{D^{(1)}_X, ..., D^{(C)}_X\} $) 
    \STATE
    \REPEAT
    \STATE ($G^{(A)}, H^{(A)}$) $\gets$ ($\emptyset, \emptyset$)
    \FOR{$c \gets 1 \text{ to } C$}
    \STATE $A$ transmits $f^{(A)}_n$ to Client: $f^{(c_i)}_n$ $\gets$ $f^{(A)}_n$
    \STATE $\mathbf{\Tilde{X}^{(c_i)}}$ $\gets$ \textit{sampling\_technique}( $\lambda, S$)
    \STATE $c_i$ Generate Predictions: $\hat{y}^{(c_i)}_n = f^{(c_i)}_n (D^{c_i}_{\Tilde{X}^{(c_i)}})$
    \STATE $c_i$ Computes $g^{(c_i)}$ and $h^{(c_i)}$, transmit to $A$
    \STATE $G^{(A)}$ $\gets$ $G^{(A)} \cup g^{(c_i)}$
    \STATE $H^{(A)}$ $\gets$ $H^{(A)} \cup h^{(c_i)}$
    \ENDFOR
    \STATE $G^{(A)}_m, H^{(A)}_m$ $\gets$ $\textit{merge\_hist}(G^{(A)}, H^{(A)})$
    \STATE $f^{(A)}_n$ $\gets$ \textit{grow\_tree}($\Tilde{D}$$^{(A)}_X, G^{(A)}, H^{(A)}$)
    \UNTIL{$n = N$ or $k = K$}
    \end{algorithmic}
\end{algorithm}
Each client constructs local histogram $\Tilde{D}^{(c_i)}_X$ using specified histogram bin size $B$ and local dataset $D^{(c_i)}_X$ and sends them to the aggregator where they are merged into a single histogram $\Tilde{D}^{(A)}_X$ (Line 3-7). Then, the iterative FL process starts. For each round, we reset the aggregated parameters $ G^{(A)}, H^{(A)}$ to 0. Then, for all clients in parallel, the aggregator $A$ transmits global model $f^{(A)}_n$ to each client which is assigned to client $c_i$'s local model $f^{(c_i)}_n$ (Line 12). We sample training data using either MVS or uniform sampling (Line 13). The integration of the sampling technique comes at a crucial point before deriving the gradients and hessians. The locally sampled dataset is derived from the $S \%$ sampled training data. Predictions $\hat{y}^{(c_i)}_n$ (Line 14) are generated and we use them for computing the local gradient $g^{(c_i)}$ and hessian $h^{(c_i)}$, before transmitting them to the Aggregator (Line 15-17). After transmission, they form aggregated gradients and hessians $ G^{(A)}, H^{(A)}$ (Line 19). They are used with the aggregated histogram representation $\Tilde{D}^{(A)}_X$ to continue to grow the tree (Line 20). 
Figure \ref{fig:(METHOD)_xgboost_flow} shows the system overview of construction our federated XGBoost. For clients, histograms are computed, samples are selected and weak learners are created. For the server, gathered parameters are merged and later evaluated as a model. Final model is created when termination criteria are satisfied. Federated XGBoost implementation can be made public upon accepted publication. 

\section{Results}\label{sec:results}
We call our model F-XGB and present its performance in comparison with other federated XGBoost frameworks. We include results using MVS- and uniform sampling on federated tabular datasets. We define federated datasets as data with (1) natural keyed generation process (keys refers to unique users), and (2) distribution skew across users/devices \cite{caldas2018leaf}. We argue that it is desirable to test federated models on such datasets as it is a natural and realistic approach to evaluate FL due to non-IID properties in distributed data. We also include 3 non-federated datasets to compare our models to relevant work. Thereafter, we present \textit{FedTab}, a collection of federated tabular datasets covering many different tasks, see Table \ref{tab:my_label}. To our knowledge, no open set of federated tabular datasets have been collected.  
We use 8GB NVIDIA GeForce RTX 3080 GPU for experiments. 
See Table \ref{tab:hyperparameter_search_space} for hyperparameter search space.

\begin{table}[tb]
    \centering
    \caption{Hyperparameter space used for F-XGB experiments. }
    \begin{tabular}{l|r}
    \hline
    \textbf{Hyperparameter} & \textbf{Search Space Values} \\
    \hline
        learning Rate ($\eta)$ & $\{0.001, 0.01, 0.02, 0.05, 0.1 \}$ \\
        lambda ($\lambda$) & $\{0.001, 0.01, 0.02, 0.05, 0.1\}$ \\
        max\_depth & $\{3, 4, 5, 6, 7, 8 \}$ \\
        max\_bin & 256\\
        sampling\_fraction (\%) & \{10, 20, 30, 40, 50\}\\
        \hline
    \end{tabular}
    \label{tab:hyperparameter_search_space}
\end{table}
\subsection{Related Model Comparison}\label{subsec:(RESULTS)_comptetitive_models}
First, we compare F-XGB to PAX \cite{ong2020adaptive} on the Airline Delay dataset. \cite{ong2020adaptive} tested their model using mentioned dataset and we follow the same pre-processing steps and use 3 clients with 1000 training instances each, and 1000 for testing. We compare F-XGB using MVS to PAX on this dataset as their model is not open-source. Random and balanced setting refers to the dataset sampling used, selecting samples based on uniform distribution or balanced in which each client has the same amount of labels for all classes. From Table \ref{tab:airline_delay_comparison}, we read that F-XGB can nearly predict all samples correctly. 

\begin{table}[tb]
    \centering
    \caption{Model accuracy and AUC on Airline Delay dataset. F-XGB uses MVS and 50\% sampling fraction. }
    \begin{tabular}{c|cc|cc}
        \hline
        & \multicolumn{2}{c}{\textbf{Airline (Random)}}& \multicolumn{2}{c}{\textbf{Airline (Balanced)}}\\
        \hline
        \textbf{Model}  & \textbf{Acc} & \textbf{AUC} & \textbf{Acc} & \textbf{AUC}\\ 
        \hline
        F-XGB & 0.97 & 0.91 & 0.97 & 0.98\\ 
        PAX & 0.88 & 0.87 & 0.87 & 0.87\\
        \hline
    \end{tabular}
    \label{tab:airline_delay_comparison}
\end{table}

\begin{table}[tb]
    \centering
    \caption{F1 scores for specified datasets. F-XGB results using MVS and sampling fraction $s \in \{10, 20 ,30, 40, 50\}$\%.}
    \begin{tabular}{c|c|c|c|c}
        \hline
         & \multicolumn{2}{c}{\textbf{Credit Card}}& \multicolumn{2}{c}{\textbf{Firewall}}\\
        \hline
        \textbf{Partition} & F-XGB & FX & F-XGB & FX\\
        \hline
        Even & 0.82& 0.80& 0.85&0.77 \\
        A &0.82 & 0.78& 0.85& 0.77\\
        B & 0.83 &0.79 & 0.85&0.77 \\
        C & 0.83&0.83 & 0.84& 0.77\\
        D &0.84 & 0.84& --- & ---\\
        \hline
    \end{tabular}
    \label{tab:non_iid_xgboost_comparison}
\end{table}

\begin{table*}[!h]
    \centering
    \caption{F-XGB performance on six datasets. Mean scores across 5 runs, each run we extract top 1\% evaluation score. MVS50 stands for F-XGB using MVS and a 50\% sampling fraction. NS and U stands for no sampling and uniform, respectively. Centralized XGBoost uses the same hyperparameter space as for federated case. }
    \begin{tabular}{l|cc|cc|cc|cc|cc|cc}
    \hline
       & \multicolumn{2}{c}{\textbf{FEMNIST}} & \multicolumn{2}{c}{\textbf{Synthetic}} & \multicolumn{2}{c}{\textbf{Heart}} &\multicolumn{2}{c}{\textbf{Machine}} & \multicolumn{2}{c}{\textbf{Skin}} & \multicolumn{2}{c}{\textbf{Insurance}}\\
       \hline
        \textbf{Model}&\textbf{Acc} & \textbf{F1} & \textbf{Acc} & \textbf{F1} & \textbf{Acc} & \textbf{AUC}  & \textbf{Acc} & \textbf{AUC} & \textbf{Acc} & \textbf{AUC}  & \textbf{RMSE} & \textbf{R2}\\
        \hline
        NS100 & $0.897$& $0.892$& $0.862$&$0.683$& $0.802$&$0.885$ & $0.982$&$0.983$&$\mathbf{0.980}$ & $0.995$& $4496$ & $0.829$ \\ 
        \hline
        MVS10 & $0.902$&$0.894$ & $0.863$&$0.686$& $\mathbf{0.881}$&$\mathbf{0.922}$&$\mathbf{0.989}$&$\mathbf{0.985}$& $0.974$& $0.996$&$4788$ & $0.847$\\
        MVS20  & $0.916$ & $0.914$ & $0.858$& $0.658$& $0.871$ & $0.917$ & $0.985$ & $0.983$ & $0.975$ & $0.996$ & $\mathbf{4082}$ & $\mathbf{0.889}$\\
        MVS30 & $0.917$&$0.913$ & $0.867$& $0.684$& $0.844$&$0.910$ & $0.987$&$0.980$ &$0.970$ &$0.995$ & $4480$ &$0.858$ \\
        MVS40  & $0.920$ & $0.919$ & $0.866$ & $0.680$ & $0.823$ & $0.900$ & $0.987$ & $0.977$ & $0.979$ & $\mathbf{0.998}$ & $4896$ & $0.831$\\
        MVS50 &$\mathbf{0.935}$& $\mathbf{0.932}$& $\mathbf{0.871}$&$\mathbf{0.702}$& $0.804$&$0.880$ &$0.984$ &$0.965$ & $\mathbf{0.980}$&$\mathbf{0.998}$&  $4429$&$0.861$ \\
        \hline
        U10  & $0.889$&$0.883$ & $0.859$&$0.668$ & $0.837$& $0.915$& $0.971$&$0.965$ & $0.956$& $0.992$& $4705$& $0.851$\\
        U20  & $0.899$ & $0.887$& $0.862$ & $0.658$ & $0.834$ & $0.910$ & $0.972$ & $0.965$ & $0.964$ & $0.993$ & $4310$ & $0.882$\\
        U30 & $0.867$& $0.862$& $0.826$ &$0.663$ &$0.803$& $0.888$& $0.987$& $0.977$ & $0.965$& $0.992$& $4451$&$0.869$ \\
        U40  & $0.875$& $0.869$&$0.843$ & $0.694$ & $0.797$ & $0.880$ & $0.983$ & $0.975$ & $0.978$ &$\mathbf{0.998}$ & $4122$ & $0.856$\\
        U50 & $0.901$& $0.891$&$0.859$ &$0.695$ & $0.7824$& $0.852$& $0.982$&$0.972$ &$0.978$& $\mathbf{0.998}$& $4630$& $0.820$\\
        \hline
        \hline
        \hline
        Central & 0.930& 0.926& 0.880 & 0.721 & 0.851 & 0.880 & 0.990 & 0.986 & 0.976 & 0.979 & 4002& 0.889 \\
       \hline
    \end{tabular}
    \label{tab:flexbo_main}
\end{table*}

We thereafter compare F-XGB to the proposed federated XGBoost by \cite{jones2022federated} (denoted FX) which is based upon the work of \cite{ong2020adaptive}. Their study provides a federated XGBoost on Non-IID data, a prominent challenge in FL, and we investigate whether F-XGB can handle such data partitions. We compare the models on 2 of the datasets included in their study: (1) Credit Card and (2) Firewall datasets as they are used in many other related studies, and use their hyperparameter setting. We complete the same partitions to form non-IID datasets and show it in Table \ref{tab:non_iid_xgboost_comparison}. F-XGB outperforms FX on both datasets in (almost) all partitions, except for one in which they perform equally good. We show the best F-XGB results using MVS and sampling fraction $s \in \{10, 20 ,30, 40, 50\}$\%.  

As shown in \cite{chen2021secureboost+}, a GOSS sampling technique in a federated setting does not improve performance in terms of accuracy. It is important to highlight, as sampling in centralized learning can improve performance. They evaluate their model in a vertical FL setting which prevents a F-XGB comparison. To our knowledge, there is no other paper that has investigated either GOSS or MVS in federated XGBoost. Thus, we evaluate F-XGB on federated tabular datasets in Subsection \ref{subsec:(RESULTS)_federated_tabular_datasets}. 

\subsection{Federated Tabular Datasets}\label{subsec:(RESULTS)_federated_tabular_datasets}
We study the performance of F-XGB on 1 regression-, 2 multiclass-, and 3 binary datasets and include results from centralized XGBoost. Scores presented are mean scores over 5 runs. We initially split the dataset into 80\% training data and 20\% validation data. We train our model for 200 rounds. 

From Table \ref{tab:flexbo_main}, we read that F-XGB using MVS outperforms other variants of the model in almost all cases. On datasets Lumpy Skin Disease and Insurance Premium Prediction, the scores are the most similar. For lumpy skin classification, the scores are almost perfect for a \textit{no sampling} (NS) F-XGB, thus sampling has a marginal effect. We notice that F-XGB using MVS with 50\% sampling fraction is the best performing model on larger- and multiclass datasets. We do not see similar behavior for uniform sampling. F-XGB using MVS and 10\% sampling fraction shows better performance on smaller and binary classification datasets. For regression task, a sampling fraction of 20\% performs the best for F-XGB using MVS. 

We include other metrics than accuracy and Root Mean Squared Error (RMSE), namely F1, AUC and R2 scores to mitigate the risk of only analyzing non-representative scores for unbalanced datasets. As shown in Table \ref{tab:flexbo_main}, F-XGB using MVS  achieves good F1 and AUC scores and for Insurance Premium dataset it is almost able to explain 90\% of the variance in the target using the features (R2 score), similarly for centralized XGBoost. As seen in Table \ref{tab:my_label}, datasets like Heart Disease, Insurance Premium Prediction, and Machine Failure include only 3-4 unique splits. Thus, we use 3-4 clients where applicable. For the remaining datasets, we sample 22 clients each run due to computational limitations. 

\begin{figure*}[tb]
    \centering
    \includegraphics[width=\linewidth]{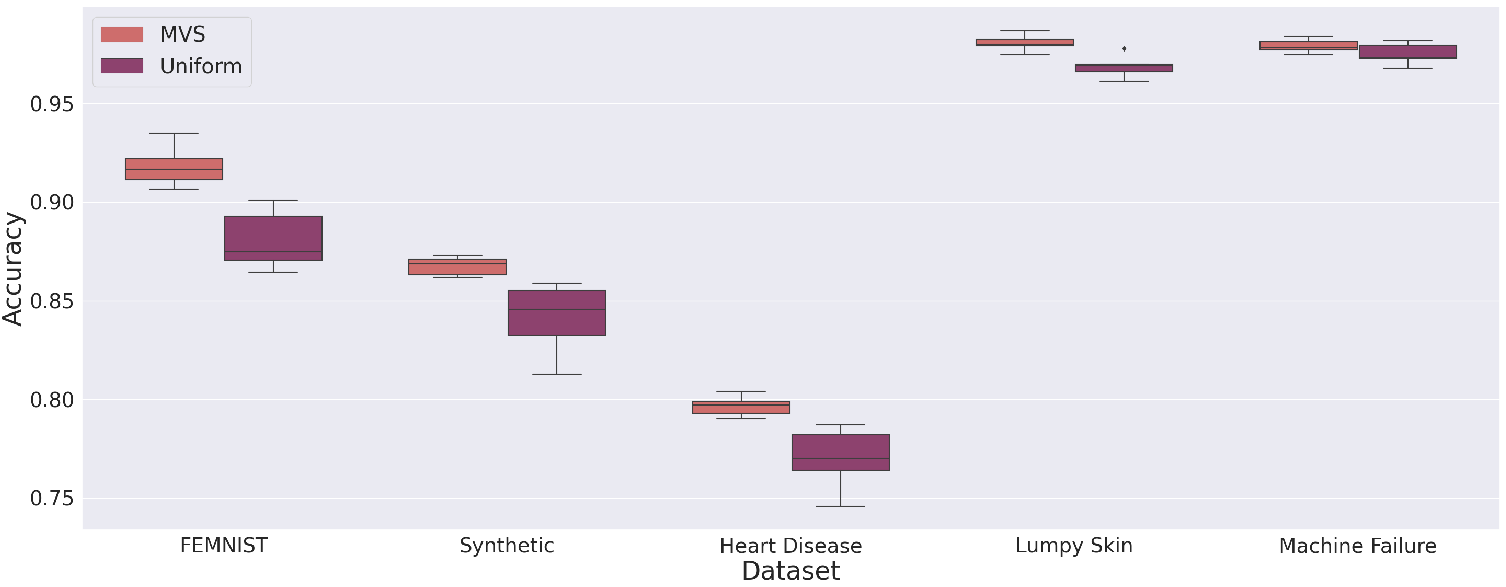}
    \caption{Top 1\% evaluation accuracy for different F-XGB sampling methods. 50\% sampling fraction is used for F-XGB on all datasets and mean scores across 5 runs}
    \label{fig:boxplot}
\end{figure*}

Interestingly, F-XGB using MVS outperforms F-XGB when no sampling is applied. Uniform sampling can increase performance but performs similarly to no-sampling in most cases. We find that F-XGB using MVS outperforms centralized XGBoost. It does so in 3 cases and perform similarly to it in the remaining cases. This is important, as the general consensus is that FL can enhance privacy and reduce communication but oftentimes comes with a loss in performance. We include the results from centralized XGBoost in Table \ref{tab:flexbo_main}, using the same hyperparameters for both federated and centralized models.

The standard deviation for all sampling methods for Insurance Premium Prediction regression task is also worth highlighting. The average standard deviation is approx. 10\% of mean value, in comparison to second largest standard deviation 2.5\% for FEMNIST. Moreover, we show in Figure \ref{fig:boxplot} how F-XGB using MVS and a sampling fraction of 50\% achieves better performance in terms of smaller standard deviation. The figure clearly illustrates how variance in predictions is reduced using F-XGB and MVS compared to uniform sampling. Uniform sampling can in certain cases perform similarly to MVS but over several runs, the performance is significantly less. For Lumpy Skin and Machine Failure dataset, the predictions over several runs are fairly similar, resulting in very small standard deviation. F-XGB using MVS does not show signs of varied predictive performance over various tasks e.g. binary- and multiclass classification, nor for small and large datasets e.g. Heart Disease and FEMNIST. The average standard deviation for MVS vs. uniform sampling on selected datasets in Figure \ref{fig:boxplot} is 0.009 and 0.013 respectively.  

\begin{figure*}[tb]
    \centering
    \includegraphics[width=18cm]{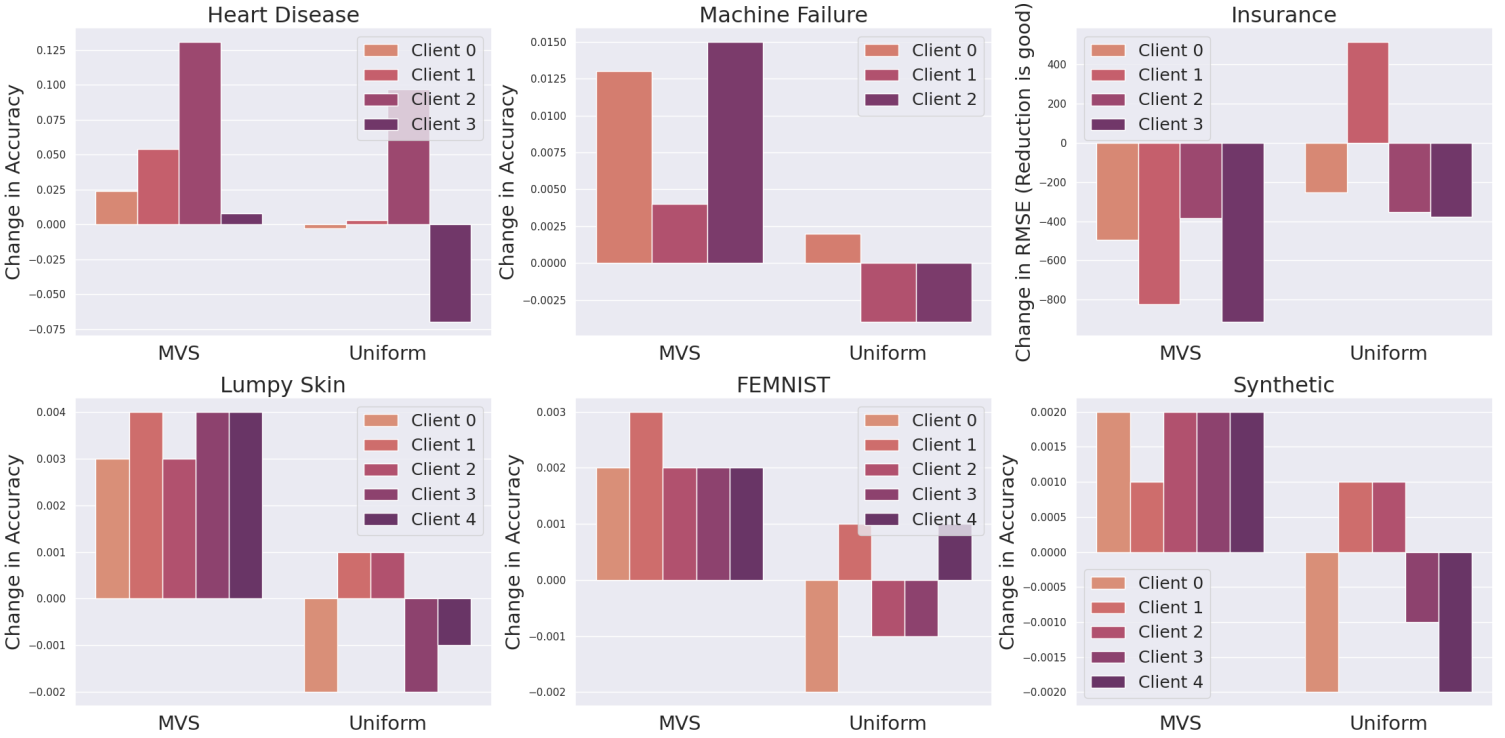}
    \caption{Change in top 1\% accuracy for local client test set vs. global evaluation set on different datasets. Scores are averaged over 5 runs. }
    \label{fig:change_in_accuracy}
\end{figure*}

Furthermore, we evaluate what the local vs. global performance in for F-XGB.  We use a train-validation-test split of 70-20-10\%, the same hyperparameter search space as described in Table \ref{tab:hyperparameter_search_space}, and compare the aggregated test scores for each client in relation with global evaluation performance. In Figure \ref{fig:change_in_accuracy}, we demonstrate F-XGB's performance using uniform sampling and MVS. Figure \ref{fig:change_in_accuracy} illustrates the change in accuracy (or RSME) on local vs. global datasets used for prediction. Importantly, as datasets Heart Disease, Machine Failure, and Insurance Premium Prediction only include 4, 3, and 4 unique splits based on their federated characteristics we only include results from 3 and 4 clients. For the other datasets, we include scores from 5 clients. We limit client selection to $\leq 5$ for illustrative purposes. 5 unique clients are randomly sampled each round. Since insurance premium prediction is a regression task, it is desirable that a local test score is lower than global evaluation since we use RMSE as metric. F-XGB using MVS outperforms uniform sampling in all classification tasks. In almost all cases, it boosts performance compared to uniform sampling for which performance mostly decreases or remains unchanged. For Lumpy Skin dataset, both sampling techniques demonstrate similar capabilities. This can be explained by that both already perform very well, almost 100\% global evaluation accuracy (see Table \ref{tab:flexbo_main}). Moreover, for Insurance Premium Prediction dataset, we see that MVS can help decrease the error while uniform sampling shows no clear signs of this behavior. For FEMNIST, Synthetic, and Lumpy Skin datasets which includes many unique users, the change in accuracy is similar among clients since we sample new clients each run. 

\section{Discussion}\label{sec:discussion}
We have presented results from our model F-XGB using sampling techniques, and both global and local performance. 
\subsection{Global Performance}\label{subsec:(DISCUSSION)_global_performance}
From the results, we read that F-XGB using MVS as a sampling technique, can increase global performance in a federated setting, which we argue is not the case for related studies that include other sampling techniques \cite{chen2021secureboost+}. The results from F-XGB using MVS are compared with no sampling and uniform sampling, see Table \ref{tab:flexbo_main}. In most cases, a sampling fraction of 50\% is recommendable to use. Presented in Table \ref{tab:flexbo_main}, MVS50 (which is F-XGB using MVS and a sampling fraction of 50\%) outperforms other model configurations. This seems reasonable as there might be a significant proportion of data that are redundant and do not have much impact on model training, thus MVS does not select them. Nevertheless, we are surprised to see that a sampling fraction of 10 or 20\% can be a more suitable choice. This could be due to data quality or many outliers in distributed datasets. Moreover, F-XGB using MVS is compared with centralized XGBoost and outperforms it in half of the studied cases. This is important as FL is often seen to improve privacy yet lower accuracy. We also show that F-XGB using MVS is able to outperform related and highly sophisticated models e.g. PAX on various datasets. F-XGB demonstrate signs of robustness on non-IID data (see Table \ref{tab:non_iid_xgboost_comparison}) which is a desirable trait in FL.  
\subsection{Local Performance}\label{subsec:(DISCUSSION)_local_performance}
In Figure \ref{fig:change_in_accuracy}, we show that F-XGB using MVS can increase performance locally when compared to global performance. In other words, in the final round of training, we evaluate both local and global model on either respective local test data or global evaluation data. From the figure, we read that F-XGB's predictive performance on local datasets can be higher than global performance, which we argue is a method of locally optimizing for data. By finding the data instances that minimize variance (i.e. MVS), F-XGB trains on the most \textit{informative} instances, avoiding redundancy. This is also argued by \cite{ibragimov2019minimal}. For the top 3 graphs in this Figure, we see that the performance improvement is quite client specific while for the remaining graphs, the effect is similar for each client because we sample 5 clients out of a larger pool. 

\subsection{Sampling in Federated Learning}\label{subsec:(DISCUSSION)_sampling_in_federated_learning}
\begin{figure}[tb]
    \centering
    \includegraphics[width=\columnwidth]{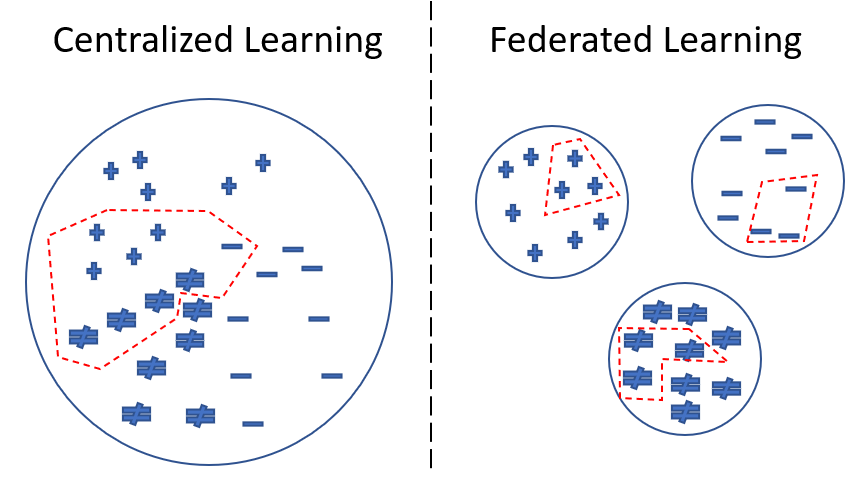}
    \caption{Different samples may be selected for training in centralized vs. federated learning. }
    \label{fig:(DISCUSSION)_central_federated_data}
\end{figure}

As presented in Section \ref{sec:related_work}, sampling has shown that performance can increase in a centralized setting, and we have now demonstrated that it can also apply to FL when using federated XGBoost and MVS. Based on the presented results, research should focus on studying sampling as a technique in FL for tabular data. As an example, will a federated model that can use an arbitrary sampling technique other than uniform sampling, select the same data instances as in a centralized setting. If so, then why? If not, what are the implications? To illustrate what we refer to Figure \ref{fig:(DISCUSSION)_central_federated_data}. Our example includes a centralized case in which 1 (-) sign is selected, 4 (+) signs and 4 ($\neq$) signs. For the contrary federated case, and arbitrary model samples $\frac{1}{3}$ of the data from each "classes", thus not the same samples for a training round. MVS seeks to minimize variance, in our case on client side. Researchers may look to explainable FL \cite{chen2022evfl} to easier visualize and give intuition to why a decentralized model in FL acts a certain way.  

\section{Conclusion}\label{sec:conclusion}
FL as a technology is gaining traction yet FL for tabular data has received less attention. XGBoost is a suitable model for such data and in a centralized setting, its performance can be improved by sampling training data when building the trees. Such sampling in a federated setting has not been concluded to improve scores e.g. accuracy, even though it is commonly used in a centralized setting. We propose a federated XGBoost that can use MVS, on federated tabular dataset that reflect the non-IID properties of natural federated data.  We study the behavior of our federated XGBoost using varying dataset tasks and sampling fraction. Our integration of Federated XGBoost with MVS significantly enhances accuracy and reduces regression error in a federated setting compared to traditional XGBoost approaches without sampling. Additionally, our federated XGBoost using MVS has similar performance to that of centralized models, outperforming them in half of the cases. Lastly, we introduce 'FedTab,' a curated collection of federated tabular datasets designed for future benchmarking studies, providing a standardized basis for evaluating federated learning methods.

\section*{Acknowledgement}
This work was partly supported by the TRANSACT project. TRANSACT (https://transact-ecsel.eu/) has received funding from the Electronic Component Systems for European Leadership Joint Under-taking under grant agreement no.  101007260.  This joint undertaking receives support from the European Union’s Horizon 2020 research and innovation programme and Austria,  Belgium,  Denmark,  Finland,  Germany, Poland, Netherlands, Norway, and Spain.

\bibliographystyle{ieeetr}
\bibliography{main.bib}

\end{document}